\documentclass[11pt]{article}

\usepackage[final]{acl}

\usepackage{times}
\usepackage{latexsym}
\usepackage[T1]{fontenc}
\usepackage[utf8]{inputenc}
\usepackage{microtype}
\usepackage{inconsolata}
\usepackage{graphicx}
\usepackage{url}
\usepackage{colortbl}
\usepackage{multirow}
\usepackage{amsmath}

\title{mdok-style at SemEval-2026 Task 10: Finetuning LLMs for Conspiracy Detection}

\author{Dominik Macko\\
Kempelen Institute of Intelligent Technologies, Bratislava, Slovakia\\
dominik.macko@kinit.sk \\}

\begin{document}
\maketitle
\begin{abstract}
SemEval-2026 Task~10 is focused on conspiracy detection. Specifically, the goal is to detect whether a Reddit comment expresses a conspiracy belief.
Our submitted mdok-style system utilizes data augmentation and self-training (to cope with a rather small amount of training data) to finetune the Qwen3-32B model for a binary text-classification task. The submitted system is very competitive, ranking in the \textbf{85th percentile} (8th out of 52 submissions). The results shown that our approach, which originated in machine-generated text detection, can be used for conspiracy detection as well.
\end{abstract}

\section{Introduction}

The Psycholinguistic Conspiracy Marker Extraction and Detection (PsyCoMark) shared task (SemEval-2026 Task~10) combines psychology and natural language processing disciplines to shed light on how conspiracy theories are expressed in social media (Reddit) conversations~\citep{samory-etal-2026-semeval}. It includes two subtasks, while we have focused only on \textbf{subtask 2} that is dedicated to binary detection of conspiracy belief in a textual comment. Specifically, the goal is to classify English Reddit comments as either conspiracy-related or not (in a Yes/No manner). The training data contains 1,715 positive samples and 2,263 negative samples. The texts have been filtered to a length of 160 to 1,000 characters.

The proposed system is heavily based on mdok (\textbf{m}achine \textbf{d}etector \textbf{o}f \textbf{K}InIT)~\citep{mdok}, a robust detector of machine-generated text winning a shared task of PAN@CLEF2025~\citep{Bevendorff2025OverviewOT}.
Our final submitted system (\textbf{mdok-style conspiracy detector})\footnote{\scriptsize\url{https://github.com/kinit-sk/mdok-style-psycomark2026}} utilizes data augmentation and self-training to cope with a rather small amount of training data. It is based on the \textbf{Qwen3-32B} model~\citep{yang2025qwen3technicalreport}, finetuned using QLoRA parameter-efficient finetuning technique (PEFT) for a binary sequence classification task.

\section{Related Works and Background}

Conspiracy detection is part of the broader field of misinformation detection. Early approaches framed the task as a text classification problem using traditional machine learning and NLP techniques, relying on lexical and stylistic features with classification models, such as SVM (support vector machines)~\citep{cortes1995support}.
Subsequent research introduced psycho-linguistic and emotional features, showing that conspiracy-related narratives exhibit distinct linguistic patterns such as heightened certainty, emotional intensity, and distrust framing. Incorporating such signals improves detection performance beyond surface-level textual representations \citep{giachanou2023detection}.
More recently, large language models (LLMs) have been applied to conspiracy detection tasks due to their strong contextual understanding. While LLMs improve generalization, they may suffer from hallucination and precision issues. Emotion-aware finetuned LLM frameworks have been proposed to integrate affective signals into transformer-based architectures, achieving improved performance over vanilla LLM baselines \citep{liu2024conspemollm}.

For the task of binary conspiracy detection within a given text, we have utilized our experience in a similar task focused on machine-generated text detection. We have transferred our best approach to this kind of a different problem.

We have already started to experiment with finetuning LLMs for a binary classification task at SemEval-2024 Task~8~\citep{spiegel-macko-2024-kinit}, where we have observed a better text-classification performance of a finetuned LLM (size of 7B parameters) in comparison to the traditionally used small pre-trained (BERT-like) models. We have further enhanced the robustness of the finetuning process~\citep{macko2025increasingrobustnessfinetunedmultilingual}, which has finally resulted in the mdok~\citep{mdok} finetuning approach. mdok has ranked 1st in both subtasks of PAN@CLEF2025~\citep{Bevendorff2025OverviewOT}, one of which was focused on robust binary text classification (specifically, the detection of machine-generated texts). In this shared task of SemEval-2026 Task~10, we utilize the mdok's obfuscation and anonymization of training data as data augmentation techniques and combine them with self-training to cope with a small amount of labeled data.

\textbf{Self-training}~\citep{AMINI2025128904} is an iterative semi-supervised learning technique, where a model acts as its own teacher to leverage unlabeled data. Firstly, a so-called teacher model is trained on a small amount of data with available ``golden'' labels (often labeled by humans). The weakly trained model is then used to predict ``silver'' labels (a.k.a., weak labels or pseudo labels, i.e., they may contain erroneously labeled samples) using any unlabeled data. Since such labels might be inaccurate, usually only the most confident predictions are used to retrain the model. This process can repeat, allowing to continuously expand the training set. However, the errors from the first iteration of silver labels can propagate throughout the learning.

\section{System Overview}

As mentioned above, the PsyCoMark shared task contains rather small amount of train data (at least for bigger LLMs). To cope with the problem we are proposing a combination of data-augmentation techniques as well as using self-training to incrementally label the provided unlabeled dev and test sets.

\subsection{Data Augmentation}

In our mdok-style approach for binary conspiracy detection, we use four data-augmentation techniques described below. To use each technique, we have copied the training texts that have been afterwards modified by the corresponding technique. In order to reduce prevalence of augmented texts (in comparison to original texts), we have used only 10\% out of each technique to combine into the training. De-duplication of the final train set removes the redundant texts (i.e., unmodified copies).

\paragraph{Anonymization.}
The orignal PsyCoMark data have already replaced URLs in the texts by a tag of [URL]. Similarly, we have used a text-prepocessing procedure (available in the original mdok) to replace the regex-identified email addresses, user mentions, and phone numbers by the tags of [EMAIL], [USER], [PHONE]. In comparison to the original mdok, we are not using this anonymization procedure for preprocessing all the texts, but for modification of copied data for data augmentation.

\paragraph{Lower-casing and Upper-casing.}
We consider conspiracy beliefs to be invariant on the casing of the text (especially in social media, often containing informal style). Therefore, to make the detection more focused on meaning rather than on the visual form, we integrate both, lowercase and uppercase copies of the original texts. This augmentation not only increases the number of training samples, but it also makes the detection case insensitive.

\paragraph{Homoglyphication.}
Homoglyph attacks (swapping visually similar characters of different scripts) are quite successful in confusing text classifiers (dependent on internal representation of characters and words, such as in tokenization). In our previous work on machine-generated text detection~\citep{macko-etal-2024-authorship}, we have found out that this confusion effect on classifiers can be effectively dealt with by inclusion of homoglyphied samples during training. As an data augmentation, we have used such homoglyphication of copied texts, increasing the train-set size while also making the conspiracy detector more robust.

\subsection{Self-Training}

In the first round, the selected LLM has been trained purely on the provided train set. Afterwards, it provided predictions of labels of the dev and test sets, while for each prediction we have dumped also a probability of the positive class (``Yes''). Based on such probabilities, we have kept as silver labels only those that had very high probability ($\geq$0.99) as positive and those that had very low probability ($\leq$0.01) as negative (in order to minimize propagation of errors). Such silver-labeled samples have been afterwards combined with the training data for retraining the detector.

\subsection{Finetuning Process}

Since the provided dev split of the data does not contain labels, we have used a hold-out training data (100 samples per class) for validation during finetuning. Inclusion of 10\% of data from each data-augmentation technique and subsequent de-duplication resulted in 2,126 negative and 1,517 positive samples for training. After inclusion of silver-labeled dev and test samples, the training set contained 2,575 negative and 1,881 positive samples.

For finetuning, we have used QLoRA~\citep{NEURIPS2023_1feb8787} parameter-efficient finetuning technique (PEFT) (4-bit quantization). As a framework, we have used the transformers\footnote{\url{https://github.com/huggingface/transformers}} python library. We have used paged adamw optimizer with cosine learning rate of 2e-5 and a warmup ratio of 0.03. We have used a batch size of 1 sample without gradient accumulation and validation each 100 steps. The training process has run a single epoch and the final checkpoint selection was based on the best Macro F1 score.

In the footnote of the first page, we have provided a link to Github repository, where we have published the full source code for replication purposes. To install the dependencies, just install the corresponding conda environment of the IMGTB\footnote{\url{https://github.com/kinit-sk/IMGTB}} framework~\citep{spiegel-macko-2024-imgtb} and update the transformers library (for the support of the newest models).

\subsection{Base Model Selection}

We have exploited the mdok-style efficient finetuning process to train various LLMs, up to the size of 32B parameters.
We have focused on Qwen3~\citep{yang2025qwen3technicalreport} and Gemma-3~\citep{gemmateam2025gemma3technicalreport} families. Based on the results in development phase (without the self-training component) using dev-set evaluation by the organizers, we have finally selected Qwen3-32B, as the best performing model for the task.

\section{Experimental Setup}

We have used only the officially provided data in the PsyCoMark shared task, which have been augmented as described in Section~3.1. We have trained the detectors without self-training and with self-training (using the silver labels predicted by Qwen3-32B). The evaluation is done only using the official shared task Codabench site\footnote{\url{https://www.codabench.org/competitions/10749}}, by the organizers, since the ground truth is not release yet. Upon the release, the error analysis might be executed.

The official metric in subtask 2 is the macro-averaged F1-score (representing a harmonic mean of precision and recall while invariant to class imbalance).

\section{Results}

The results of various system alternatives are provided in Table~\ref{tab:performance}. The results of the compared alternatives are very close to each other. The submitted highlighted system is using self-training (denoted \_ST) and classification threshold moved to 0.7 of positive-class probability (denoted \_th0.7). However, even the smallest tested DeBERTa model (<0.5B parameters) achieved 0.75 of Macro F1 score, offering a better tradeoff of performance for the cost.

\begin{table}[!t]
\centering
\resizebox{0.7\linewidth}{!}{
%\addtolength{\tabcolsep}{-2pt}
\begin{tabular}{r|c}
\hline
\bfseries Detector & \bfseries Macro F1 \\
\hline
\cellcolor{lightgray}\textbf{Qwen3-32B\_ST\_th0.7} & \cellcolor{lightgray}\textbf{0.78} \\
Qwen3-32B\_ST & 0.77 \\
Qwen3-32B\_th0.7 & 0.77 \\
Qwen3-32B & 0.76 \\
DeBERTa-Large & 0.75 \\
Qwen3-14B-Base & 0.75 \\
Gemma-3-1B-PT\_ST & 0.75 \\
Qwen3-4B-Base & 0.75 \\
Gemma-3-12B-PT & 0.74 \\
Gemma-3-1B-PT & 0.73 \\
Qwen3-4B-Base\_ST & 0.72 \\
\hline
random baseline & 0.50 \\
\hline
\end{tabular}
}
%\vspace{-2mm}
\caption{The performance of the various system alternatives using the official test set for the PsyCoMark subtask 2.}
\label{tab:performance}
%\vspace{-3mm}
\end{table}

The unofficial ranking (provided currently in the Codabench shared task site, some teams might be disqualified yet if not submitting system-description papers) of the submitted system is provided in Table~\ref{tab:rank}. As illustrated, the mdok-style system ranked competitively, in \textbf{the top 20\%} of the submissions (~85th percentile).

\begin{table}[!t]
\centering
\resizebox{0.75\linewidth}{!}{
%\addtolength{\tabcolsep}{-2pt}
\begin{tabular}{r|c|c}
\hline
\bfseries Rank & \bfseries Team & \bfseries Macro F1 \\
\hline
      1    & NJUST\_KMG &  0.89  \\
      2    & AGAI &  0.87  \\
      3    & jia57 &  0.86  \\
      4    & baishanxiaoqi &  0.8  \\
      5    & CSECU-DSG &  0.8  \\
      6  & joccerrillo &  0.79  \\
      7  & qinchihongye &  0.79  \\
      \cellcolor{lightgray}\bfseries 8  & \cellcolor{lightgray}\bfseries mdok-style &  \cellcolor{lightgray}\bfseries 0.78  \\
      9  & dangphuduy &  0.78  \\
      10  & shubham\_bits &  0.78  \\
      11  & rziaei &  0.77  \\
      12  & dmarhoef &  0.77  \\
      13  & jorgegomezn &  0.77  \\
      14  & srikarkashyap &  0.77  \\
      15  & shahmir2002 &  0.76  \\
      16  & CuriosAI &  0.76  \\
      17  & dengkuihou &  0.76  \\
      18  & joesrwt &  0.76  \\
      19  & Tilde &  0.76  \\
      20  & Macaroni &  0.76  \\
      21  & davidinfotec &  0.75  \\
      22  & psy\_detectives &  0.75  \\
      23  & ishaank &  0.75  \\
      24  & panos\_span &  0.75  \\
      25  & autist &  0.74  \\
      26  & Macaroni &  0.74  \\
      27  & wangkongqiang &  0.74  \\
      28  & YNU-HPCC &  0.74  \\
      29  & TM &  0.74  \\
      30  & Sarang &  0.73  \\
      31  & hidetsune &  0.73  \\
      32  & CCNU &  0.73  \\
      33  & pc0907 &  0.73  \\
      34  & pannndoo &  0.73  \\
      35  & zhangpeng &  0.73  \\
      36  & lamiaa &  0.72  \\
      37  & stangerine &  0.72  \\
      38  & 123xxx &  0.72  \\
      39  & 777lily &  0.72  \\
      40  & yiyu12 &  0.72  \\
      41  & zzz666 &  0.72  \\
      42  & bruhh &  0.72  \\
      43  & wagetl &  0.72  \\
      44  & civilwen &  0.72  \\
      45  & xlxxlx &  0.72  \\
      46  & lemontr1 &  0.72  \\
      47  & samu721 &  0.72  \\
      48  & emmasleghel &  0.72  \\
      49  & hritav\_1896 &  0.71  \\
      50  & samorymatest &  0.71  \\
      51  & faozia\_fariha &  0.59  \\
      52  & bpiper02 &  0.41  \\
\hline
\end{tabular}
}
%\vspace{-2mm}
\caption{The unofficial ranking of the submitted system for the PsyCoMark subtask 2.}
\label{tab:rank}
%\vspace{-3mm}
\end{table}

\section{Conclusion}

Our participation in the shared task provided multiple insights. Our mdok approach, which has been developed for machine-generated text detection, has been successfully transferred to conspiracy detection task. We have compared multiple system alternatives, out of which the submitted system using data augmentation and self-training based on Qwen3-32B offers the best perfomance, ranking competitively among all the submissions. However, tradeoff between detection performance and computation costs must be considered, since the DeBERTa model achieved only slightly lower performance.

\section*{Limitations}

We have explored only small set of base language models. Others could be better performing. Since only the English texts have been included in the shared task, the generalization to other languages is unevaluated. We have limited the set of texts for finetuning only to the official data of the shared task. Other publicly available datasets could be used for training as well.

\section*{Acknowledgments}
This work was supported by the EU NextGenerationEU through the Recovery and Resilience Plan for Slovakia under the project No. 09I01-03-V04-00068.

\textbf{Computational resources}. We acknowledge EuroHPC Joint Undertaking for awarding us access to Leonardo at CINECA, Italy.

\bibliography{custom, anthology}

@inproceedings{samory-etal-2026-semeval,
    title = "{S}em{E}val-2026 Task 10: {P}sy{C}o{M}ark -- Psycholinguistic Conspiracy Marker Extraction and Detection",
    author = "Samory, Mattia  and
        Soldner, Felix  and
        Batzdorfer, Veronika",
    booktitle = "Proceedings of the 20th International Workshop on Semantic Evaluation (SemEval-2026)",
    year = "2026",
}

@article{cortes1995support,
  title={Support-vector networks},
  author={Cortes, Corinna and Vapnik, Vladimir},
  journal={Machine learning},
  volume={20},
  number={3},
  pages={273--297},
  year={1995},
  publisher={Springer}
}

@article{giachanou2023detection,
  title={Detection of conspiracy propagators using psycho-linguistic characteristics},
  author={Giachanou, Anastasia and Ghanem, Bilal and Rosso, Paolo},
  journal={Journal of Information Science},
  volume={49},
  number={1},
  pages={3--17},
  year={2023},
  publisher={SAGE Publications Sage UK: London, England}
}

@article{liu2024conspemollm,
  title={Conspemollm: Conspiracy theory detection using an emotion-based large language model},
  author={Liu, Zhiwei and Liu, Boyang and Thompson, Paul and Yang, Kailai and Ananiadou, Sophia},
  journal={arXiv preprint arXiv:2403.06765},
  year={2024}
}

@article{AMINI2025128904,
title = {Self-training: A survey},
journal = {Neurocomputing},
volume = {616},
pages = {128904},
year = {2025},
issn = {0925-2312},
doi = {https://doi.org/10.1016/j.neucom.2024.128904},
url = {https://www.sciencedirect.com/science/article/pii/S0925231224016758},
author = {Massih-Reza Amini and Vasilii Feofanov and Loïc Pauletto and Liès Hadjadj and Émilie Devijver and Yury Maximov},
}

@misc{macko2025increasingrobustnessfinetunedmultilingual,
      title={Increasing the Robustness of the Fine-tuned Multilingual Machine-Generated Text Detectors}, 
      author={Dominik Macko and Robert Moro and Ivan Srba},
      year={2025},
      eprint={2503.15128},
      archivePrefix={arXiv},
      primaryClass={cs.CL},
      url={https://arxiv.org/abs/2503.15128}, 
}

@inproceedings{Bevendorff2025OverviewOT,
  title={Overview of the "Voight-Kampff" Generative {AI} Authorship Verification Task at {PAN} and {ELOQUENT} 2025},
  author={Janek Bevendorff and Yuxia Wang and Jussi Karlgren and Matti Wiegmann and Maik Fr{\"o}be and Akim Tsivgun and Jinyan Su and Zhuohan Xie and Mervat T. Abassy and Jonibek Mansurov and Rui Xing and Minh Ngoc Ta and Kareem Ashraf Elozeiri and Tianle Gu and Raj Vardhan Tomar and Jiahui Geng and Ekaterina Artemova and Artem Shelmanov and Nizar Habash and Efstathios Stamatatos and Iryna Gurevych and Preslav Nakov and Martin Potthast and Benno Stein},
  booktitle = {Working Notes of {CLEF} 2025 -- Conference and Labs of the Evaluation Forum, {CEUR-WS.org}},
  year = 2025,
  url = {https://ceur-ws.org/Vol-4038/paper_277.pdf}
}

@inproceedings{mdok,
    author = {Dominik Macko},
    title = {mdok of {KInIT}: Robustly Fine-tuned {LLM} for Binary and Multiclass {AI}-Generated Text Detection},
    booktitle = {Working Notes of {CLEF} 2025 -- Conference and Labs of the Evaluation Forum, {CEUR-WS.org}},
    year = 2025,
    url = {https://ceur-ws.org/Vol-4038/paper_307.pdf}
}

@inproceedings{NEURIPS2023_1feb8787,
 author = {Dettmers, Tim and Pagnoni, Artidoro and Holtzman, Ari and Zettlemoyer, Luke},
 booktitle = {Advances in Neural Information Processing Systems},
 editor = {A. Oh and T. Naumann and A. Globerson and K. Saenko and M. Hardt and S. Levine},
 pages = {10088--10115},
 publisher = {Curran Associates, Inc.},
 title = {{QLoRA}: Efficient Finetuning of Quantized {LLMs}},
 url = {https://proceedings.neurips.cc/paper_files/paper/2023/file/1feb87871436031bdc0f2beaa62a049b-Paper-Conference.pdf},
 volume = {36},
 year = {2023}
}

@misc{gemmateam2025gemma3technicalreport,
      title={Gemma 3 Technical Report}, 
      author={Gemma Team and Aishwarya Kamath and Johan Ferret and Shreya Pathak and Nino Vieillard and Ramona Merhej and Sarah Perrin and Tatiana Matejovicova and Alexandre Ramé and Morgane Rivière and Louis Rouillard and Thomas Mesnard and Geoffrey Cideron and Jean-bastien Grill and Sabela Ramos and Edouard Yvinec and Michelle Casbon and Etienne Pot and Ivo Penchev and Gaël Liu and Francesco Visin and Kathleen Kenealy and Lucas Beyer and Xiaohai Zhai and Anton Tsitsulin and Robert Busa-Fekete and Alex Feng and Noveen Sachdeva and Benjamin Coleman and Yi Gao and Basil Mustafa and Iain Barr and Emilio Parisotto and David Tian and Matan Eyal and Colin Cherry and Jan-Thorsten Peter and Danila Sinopalnikov and Surya Bhupatiraju and Rishabh Agarwal and Mehran Kazemi and Dan Malkin and Ravin Kumar and David Vilar and Idan Brusilovsky and Jiaming Luo and Andreas Steiner and Abe Friesen and Abhanshu Sharma and Abheesht Sharma and Adi Mayrav Gilady and Adrian Goedeckemeyer and Alaa Saade and Alex Feng and Alexander Kolesnikov and Alexei Bendebury and Alvin Abdagic and Amit Vadi and András György and André Susano Pinto and Anil Das and Ankur Bapna and Antoine Miech and Antoine Yang and Antonia Paterson and Ashish Shenoy and Ayan Chakrabarti and Bilal Piot and Bo Wu and Bobak Shahriari and Bryce Petrini and Charlie Chen and Charline Le Lan and Christopher A. Choquette-Choo and CJ Carey and Cormac Brick and Daniel Deutsch and Danielle Eisenbud and Dee Cattle and Derek Cheng and Dimitris Paparas and Divyashree Shivakumar Sreepathihalli and Doug Reid and Dustin Tran and Dustin Zelle and Eric Noland and Erwin Huizenga and Eugene Kharitonov and Frederick Liu and Gagik Amirkhanyan and Glenn Cameron and Hadi Hashemi and Hanna Klimczak-Plucińska and Harman Singh and Harsh Mehta and Harshal Tushar Lehri and Hussein Hazimeh and Ian Ballantyne and Idan Szpektor and Ivan Nardini and Jean Pouget-Abadie and Jetha Chan and Joe Stanton and John Wieting and Jonathan Lai and Jordi Orbay and Joseph Fernandez and Josh Newlan and Ju-yeong Ji and Jyotinder Singh and Kat Black and Kathy Yu and Kevin Hui and Kiran Vodrahalli and Klaus Greff and Linhai Qiu and Marcella Valentine and Marina Coelho and Marvin Ritter and Matt Hoffman and Matthew Watson and Mayank Chaturvedi and Michael Moynihan and Min Ma and Nabila Babar and Natasha Noy and Nathan Byrd and Nick Roy and Nikola Momchev and Nilay Chauhan and Noveen Sachdeva and Oskar Bunyan and Pankil Botarda and Paul Caron and Paul Kishan Rubenstein and Phil Culliton and Philipp Schmid and Pier Giuseppe Sessa and Pingmei Xu and Piotr Stanczyk and Pouya Tafti and Rakesh Shivanna and Renjie Wu and Renke Pan and Reza Rokni and Rob Willoughby and Rohith Vallu and Ryan Mullins and Sammy Jerome and Sara Smoot and Sertan Girgin and Shariq Iqbal and Shashir Reddy and Shruti Sheth and Siim Põder and Sijal Bhatnagar and Sindhu Raghuram Panyam and Sivan Eiger and Susan Zhang and Tianqi Liu and Trevor Yacovone and Tyler Liechty and Uday Kalra and Utku Evci and Vedant Misra and Vincent Roseberry and Vlad Feinberg and Vlad Kolesnikov and Woohyun Han and Woosuk Kwon and Xi Chen and Yinlam Chow and Yuvein Zhu and Zichuan Wei and Zoltan Egyed and Victor Cotruta and Minh Giang and Phoebe Kirk and Anand Rao and Kat Black and Nabila Babar and Jessica Lo and Erica Moreira and Luiz Gustavo Martins and Omar Sanseviero and Lucas Gonzalez and Zach Gleicher and Tris Warkentin and Vahab Mirrokni and Evan Senter and Eli Collins and Joelle Barral and Zoubin Ghahramani and Raia Hadsell and Yossi Matias and D. Sculley and Slav Petrov and Noah Fiedel and Noam Shazeer and Oriol Vinyals and Jeff Dean and Demis Hassabis and Koray Kavukcuoglu and Clement Farabet and Elena Buchatskaya and Jean-Baptiste Alayrac and Rohan Anil and Dmitry and Lepikhin and Sebastian Borgeaud and Olivier Bachem and Armand Joulin and Alek Andreev and Cassidy Hardin and Robert Dadashi and Léonard Hussenot},
      year={2025},
      eprint={2503.19786},
      archivePrefix={arXiv},
      primaryClass={cs.CL},
      url={https://arxiv.org/abs/2503.19786}, 
}

@misc{yang2025qwen3technicalreport,
      title={Qwen3 Technical Report}, 
      author={An Yang and Anfeng Li and Baosong Yang and Beichen Zhang and Binyuan Hui and Bo Zheng and Bowen Yu and Chang Gao and Chengen Huang and Chenxu Lv and Chujie Zheng and Dayiheng Liu and Fan Zhou and Fei Huang and Feng Hu and Hao Ge and Haoran Wei and Huan Lin and Jialong Tang and Jian Yang and Jianhong Tu and Jianwei Zhang and Jianxin Yang and Jiaxi Yang and Jing Zhou and Jingren Zhou and Junyang Lin and Kai Dang and Keqin Bao and Kexin Yang and Le Yu and Lianghao Deng and Mei Li and Mingfeng Xue and Mingze Li and Pei Zhang and Peng Wang and Qin Zhu and Rui Men and Ruize Gao and Shixuan Liu and Shuang Luo and Tianhao Li and Tianyi Tang and Wenbiao Yin and Xingzhang Ren and Xinyu Wang and Xinyu Zhang and Xuancheng Ren and Yang Fan and Yang Su and Yichang Zhang and Yinger Zhang and Yu Wan and Yuqiong Liu and Zekun Wang and Zeyu Cui and Zhenru Zhang and Zhipeng Zhou and Zihan Qiu},
      year={2025},
      eprint={2505.09388},
      archivePrefix={arXiv},
      primaryClass={cs.CL},
      url={https://arxiv.org/abs/2505.09388}, 
}

\appendix

\section{Computational Resources}
\label{sec:A}

For experiments regarding detector training (model finetuning) and inference, we have used $1\times$ NVIDIA A100 64GB GPU, consumed approximately 100 GPU hours. Analysis has been done without the GPU acceleration.

\end{document}